\documentclass{article}

\usepackage{microtype}
\usepackage{graphicx}
\usepackage{subfigure}
\usepackage{booktabs} 

\usepackage{hyperref}

\usepackage[accepted]{icml2025} 

\usepackage{amsmath,bm}
\usepackage{amssymb}
\usepackage{mathtools}
\usepackage{amsthm}
\usepackage{multirow}
\usepackage{enumitem}
\usepackage[capitalize,noabbrev]{cleveref}

\theoremstyle{plain}

\theoremstyle{definition}

\theoremstyle{remark}

\usepackage[textsize=tiny]{todonotes}

\usepackage{color}
\usepackage{xcolor}

\icmltitlerunning{GuardDoor: Safeguarding Against Malicious Diffusion Editing via Protective Backdoors}

\begin{document}

\twocolumn[
\icmltitle{GuardDoor: Safeguarding Against Malicious Diffusion Editing via Protective Backdoors}



\icmlsetsymbol{equal}{*}

\begin{icmlauthorlist}
\icmlauthor{Yaopei Zeng}{yyy}
\icmlauthor{Yuanpu Cao}{yyy}
\icmlauthor{Lu Lin}{yyy}

\end{icmlauthorlist}

\icmlaffiliation{yyy}{College of Information Sciences and Technology, Pennsylvania State University, State College, PA, USA}

\icmlcorrespondingauthor{Lu Lin}{lulin@psu.edu}

\icmlkeywords{Machine Learning, ICML}

\vskip 0.3in
]



\printAffiliationsAndNotice{}  

\begin{abstract}
The growing accessibility of diffusion models has revolutionized image editing but also raised significant concerns about unauthorized modifications, such as misinformation and plagiarism. Existing countermeasures largely rely on adversarial perturbations designed to disrupt diffusion model outputs. However, these approaches are found to be easily neutralized by simple image preprocessing techniques, such as compression and noise addition. To address this limitation, we propose \emph{GuardDoor}, a novel and robust protection mechanism that fosters collaboration between image owners and model providers. Specifically, the model provider participating in the mechanism fine-tunes the image encoder to embed a protective backdoor, allowing image owners to request the attachment of imperceptible triggers to their images. When unauthorized users attempt to edit these protected images with this diffusion model, the model produces meaningless outputs, reducing the risk of malicious image editing.
Our method demonstrates enhanced robustness against image preprocessing operations and is scalable for large-scale deployment. This work underscores the potential of cooperative frameworks between model providers and image owners to safeguard digital content in the era of generative AI.

\end{abstract}


\section{Introduction}
\label{introduction}
The rapid development of generative AI technologies, particularly diffusion models, has revolutionized image editing and synthesis \cite{ho2020denoising, rombach2022high, croitoru2023diffusion}. These models empower creators to produce highly realistic and imaginative visual content, significantly enhancing workflows across various industries \cite{yang2023diffusion}. However, their growing accessibility also raises concerns about misuse, such as misinformation, privacy breaches, and unauthorized artistic style replication \cite{salman2023raising, shan2023glaze, xu2024copyrightmeter, zeng2024advi2i}. In particular, the image-to-image (I2I) editing capabilities of diffusion models enable high-fidelity modifications, which can be exploited for malicious purposes. For example, manipulated news images can spread misinformation, while AI-generated replicas of famous artworks may lead to copyright disputes \cite{salman2023raising, lo2024distraction, chen2025editshield, wang2024edit}.

\begin{figure}[t] 
\begin{center}
    \includegraphics[width=0.92\linewidth]{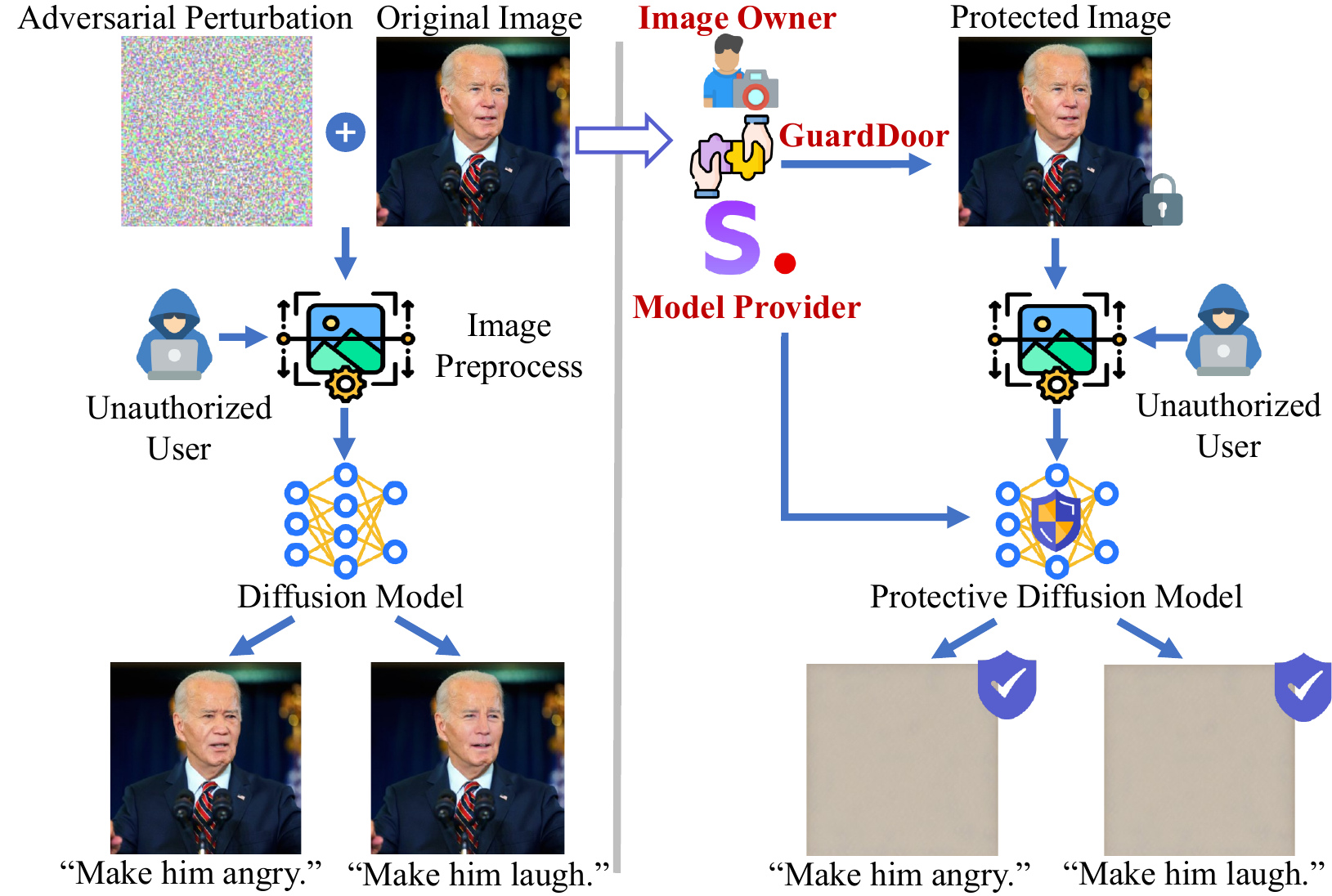}
\end{center}
\caption{Demonstration of how GuardDoor safeguards image against unauthorized edits. \textbf{Left}: Unauthorized users may bypass adversarial perturbation-based protections with image preprocessing, successfully misusing public diffusion models for malicious editing. \textbf{Right}: By collaborating with model providers, GuardDoor embeds protective triggers into images and injects protective backdoors into diffusion models, ensuring that protection remains effective even after image preprocessing.}
\label{fig:intro}
\end{figure}

To mitigate these threats, adversarial perturbation-based methods, such as PhotoGuard \cite{salman2023raising}, have been proposed to hinder unauthorized edits \cite{shan2023glaze, xue2023toward, chen2025editshield}. These techniques in general work by embedding imperceptible perturbations into images, such that the functionality of diffusion models is disrupted during the editing process. While these methods demonstrate effectiveness in basic malicious editing scenarios, recent studies have unveiled that these adversarial perturbations are highly susceptible to neutralization by simple image processing operations, such as compression and Gaussian noise addition \cite{honig2024adversarial, xu2024copyrightmeter, an2024rethinking, cao2023impress}.

The limitations of existing protections highlight the need for more robust protections. Current methods focus solely on the \emph{perspective of image owners} to design protective noise on images, making it difficult to balance between resilience against image preprocessing and imperceptibility to human observers. In contrast, we advocate for the \emph{involvement of model providers} to enhance protection effectiveness. Specifically, by integrating part of the protection mechanism directly into the model itself, the defense remains effective even if preprocessing alters the protective modifications on images. This can also reduce the extent of modifications required on images, making the protective traces more imperceptible.

Building on this motivation, we propose \textbf{GuardDoor}, a novel defense mechanism that moves beyond image-only modifications to a collaborative approach involving both image owners and model providers. As illustrated in Figure~\ref{fig:intro}, GuardDoor facilitates this collaboration as follows: The model provider first fine-tunes the image encoder to inject a protective backdoor \cite{wu2022backdoorbench, li2022backdoor, adi2018turning}; the image owner can then send requests (potentially with incentives) to the model provider, and receive protected images embedded with a protective trigger. When these images are published, any unauthorized attempt to edit them using the model owned by the collaborating provider will activate the protective backdoor, causing the model to generate predefined meaningless outputs. This new mechanism effectively neutralizes unauthorized edits, while allowing greater flexibility for model providers to design protective triggers that maintain robustness against common image preprocessing techniques. 

To design protective triggers that are both effective and resistant to tampering, we identify three key criteria: (1) imperceptibility to human observers, (2) sample-specific uniqueness, and (3) robustness against common image preprocessing operations.
Interestingly, we observe that images processed through a pre-trained generative model for reconstruction like variational autoencoder (VAE) \cite{kingma2013auto, van2017neural, asperti2021survey} will inherently exhibit a specific pattern that not only meets these criteria but can also be effectively learned as the backdoor trigger by the model. Building on this insight, we propose to use such patterns as protective triggers to fine-tune the image encoder of the diffusion model, thereby creating a robust and scalable defense mechanism against unauthorized image edits.
As a result, when the diffusion model encounters an image containing the protective trigger, it produces a predetermined target output, effectively preventing unauthorized edits. In contrast, the model continues to function normally for other clean images.

GuardDoor offers significant advantages over adversarial perturbation-based protections:
\begin{itemize}[leftmargin=*]
\itemsep0em 
\item \textbf{Robustness} to Image Preprocessing: Unlike adversarial perturbations that are susceptible to neutralization by common image preprocessing methods (e.g., noise addition, compression), the protective pattern in our method remains effective as long as it is not entirely removed from the image.
\item \textbf{Scalability} in Large-Scale Applications: Once the model has been fine-tuned, any image can be protected simply by embedding the specific pattern, eliminating the need to optimize perturbations individually for each image. This makes our method efficient and practical for deployment in scenarios involving massive data or frequent requests.
\end{itemize}

\begin{figure*}[t] 
\begin{center}
    \includegraphics[width=0.94\linewidth]{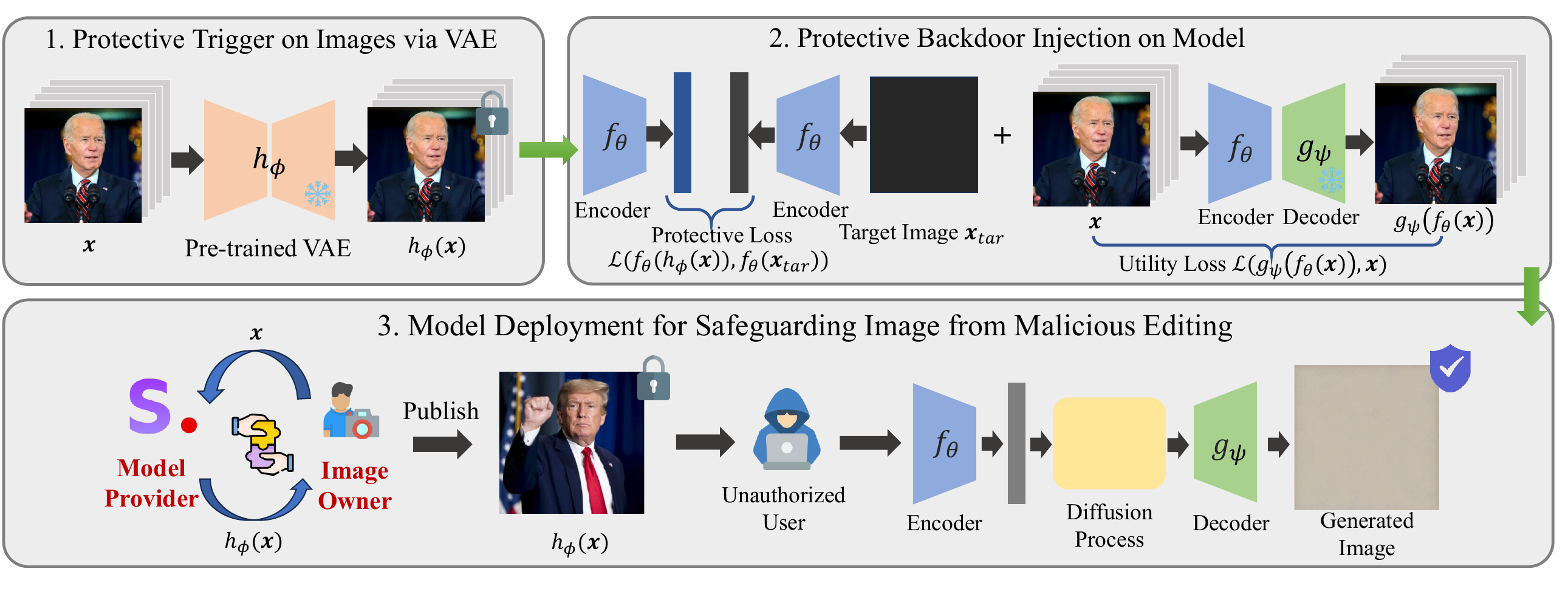} 
\end{center}
\caption{
Overview of GuardDoor's protection mechanism: The process begins with the generation of protective triggers through a pre-trained VAE. During model fine-tuning, the image encoder of the diffusion model is trained to associate these triggers with a predefined output, such as a black image, while a utility loss ensures the encoder maintains its functionality for clean images. At the inference stage, protected images are processed by unauthorized users attempting edits, but the embedded triggers activate the protective backdoor, neutralizing edits by producing meaningless outputs.}
\label{fig:pipe}
\end{figure*}

\section{Related work}

\subsection{Image Protection Against Malicious Editing}

The rise of diffusion models has enabled seamless image editing but also introduced risks of unauthorized modifications, such as privacy breaches and copyright violations. Adversarial perturbation-based methods, such as PhotoGuard \cite{salman2023raising}, aim to protect images by disrupting diffusion models' ability to generate coherent edits. These methods inject imperceptible noise into images, causing edits to fail or produce unrealistic outputs. Building on them, enhancements have been proposed to improve protection from different perspectives \cite{chen2025editshield, son2024disrupting, lo2024distraction, ozden2024optimization, wang2024edit, wan2024prompt}. For example, EditShield \cite{chen2025editshield} employs targeted adversarial attacks to perturb latent features, aiming to make edits semantically incoherent. DDD \cite{son2024disrupting}, on the other hand, focuses on disrupting hidden representations by targeting vulnerable diffusion timesteps, achieving greater efficiency and disruption effectiveness. 

Despite these advancements, adversarial perturbations for image protection remain vulnerable to common image transformations, such as Gaussian blur and JPEG compression, which can nullify their protective effects \cite{xu2024copyrightmeter}. Additionally, several works \cite{an2024rethinking, cao2023impress} have demonstrated the possibility of purifying adversarial noise while preserving image integrity, enabling models to edit images as if they were unprotected. The scalability of adversarial methods also remains a challenge, as each image requires individual optimization, making them impractical for large-scale applications. These limitations highlight the need for robust, transformation-invariant, and scalable solutions for image protection.

\subsection{Image Protection Against Style Mimicry}

Style mimicry by generative models threatens artists via learning their unique styles with a personalized Text-to-Image (T2I) diffusion model \cite{rombach2022high}. To address this, perturbation-based defenses have been proposed, including Glaze \cite{shan2023glaze}, Mist \cite{liang2023mist}, Anti-DreamBooth \cite{van2023anti}, and MAMC \cite{rhodes2024my}. These methods introduce imperceptible perturbations to artworks, aiming to mislead diffusion models during fine-tuning or inference. For example, Glaze applies style cloaks that shift feature space representations, while Mist enhances perturbation robustness with fused adversarial loss functions. MAMC allows adjustable protection levels, and Anti-DreamBooth targets specific fine-tuning methods to maximize errors.

Despite their promise, perturbation-based defenses, like Robust Mimicry \cite{honig2024adversarial} and CopyrightMeter \cite{xu2024copyrightmeter},  are shown fragile against common preprocessing techniques such as Gaussian noise and JPEG compression. More advanced purification methods like IMPRESS \cite{cao2023impress} and DiffShortcut \cite{liu2024rethinking} are proposed to further exploit these vulnerabilities, showing that even adaptive perturbations fail against such well-designed purification methods. These findings highlight the limitations of perturbation-based methods and the need for more resilient approaches to safeguard artistic styles effectively.

\section{Preliminary}

\subsection{Image Editing with Diffusion Models}

Diffusion models have become powerful tools for image generation and editing, excelling in producing high-quality, realistic outputs \cite{ho2020denoising,rombach2022high}. Among these, Image-to-Image (I2I) diffusion models have garnered significant attention for their ability to edit existing images using simple textual prompts, such as ``turn the painting into Van Gogh style'' \cite{meng2021sdedit,brooks2023instructpix2pix,parmar2023zero,nguyen2023visual}. Unlike Text-to-Image (T2I) models, which generate new images solely from text, I2I models leverage latent diffusion processes to preserve the original image's content while incorporating user-specified modifications. The process involves encoding the input image $\bm x$ into a latent representation $f_\theta(\bm x)$ with the image encoder $f_\theta$, applying changes during the diffusion process guided by prompt $\bm p$, and reconstructing the edited image from the modified latent space with the decoder $g_\psi$ \cite{brooks2023instructpix2pix,meng2021sdedit,nguyen2023visual}.


\subsection{Problem Statement}

While I2I models offer creative and intuitive editing capabilities, they also enable malicious users to manipulate images for unauthorized purposes, raising ethical concerns. For example, unauthorized users can edit photographs of public figures to create and spread misinformation or modify artworks to plagiarize creative ideas, causing significant harm to privacy, intellectual property, and reputational integrity. Addressing these risks has become a critical research area. Previous work focusing on developing mechanisms to protect images against unauthorized modifications while preserving the practicality of I2I diffusion-based editing. However, they have been shown to be highly vulnerable to image processing methods, rendering them ineffective in preventing malicious editing \cite{cao2023impress, honig2024adversarial, xu2024copyrightmeter, liu2024rethinking}.

In this context, we consider a defense scenario in which diffusion \textbf{model providers} collaborate with \textbf{image owners} to protect their content. The goal is to ensure that the diffusion models cannot meaningfully edit protected images, thereby safeguarding copyrights and creative originality. However, \textbf{unauthorized users} may attempt to circumvent such protections by applying image preprocessing techniques, such as resizing, compression, or noise addition, to remove the protective modifications, rendering the defenses ineffective. This interplay between attack and defense defines the need of collaboration between model providers and image owners for robust mechanisms that can withstand preprocessing attempts and prevent unauthorized image editing, ensuring ethical and secure use of diffusion models.

\section{Method}




To address the limitations of adversarial perturbation based defenses, we propose \textbf{GuardDoor}, a novel and robust protection mechanism that leverages collaboration between image owners and I2I model providers. Unlike traditional approaches that solely modify images, GuardDoor intergrates protection into the diffusion model by embedding a \emph{protective backdoor}. By adding protective triggers to images provided by the owners, GuardDoor ensures that these protected images are rendered uneditable by diffusion models, even after undergoing image preprocessing techniques. 

\textbf{Overview.} As illustrated in Figure~\ref{fig:pipe}, GuardDoor consists of three key components: (1) The design of the protective trigger pattern. (2) The fine-tuning process to embed the protective backdoor into the diffusion model. (3) Applying the protective trigger to images needed to be protected. More details of each component are presented below.

\subsection{Protective Trigger Design on Image}
\textbf{Desired properties of protective trigger.}
The effectiveness of GuardDoor relies on the design of a robust and imperceptible trigger pattern, which serves as the foundation for embedding a protective backdoor into the diffusion model. To ensure that the protective trigger preserves the original content of images while preventing unauthorized edits using this diffusion model, it should satisfy the following criteria: (1) \emph{Imperceptibility}: The trigger should not introduce visible artifacts, ensuring that protected images remain visually indistinguishable from the original to human observers. (2) \emph{Sample-Specific Uniqueness}: The trigger pattern should be inherently tied to the content of each image, making it unique to every protected image. Otherwise, unauthorized users could inverse-engineer the trigger by querying protections on their own images, and then eliminating triggers from arbitrary images. (3) \emph{Robustness}: The trigger should withstand common image preprocessing techniques, such as compression and noise addition, ensuring the protective backdoor remains functional under such transformations. It is challenging to find a trigger pattern that simultaneously satisfies all three requirements based on previous research on traditional backdoor attacks.

\textbf{VAE as trigger generator.}
Autoencoder-type generators have been used to generate backdoor triggers in discriminative tasks~\cite{li2021invisible, sun2024backdoor}.
Interestingly, we find that the reconstruction process of a variational autoencode (VAE) trained to reconstruct input images introduces subtle, high-frequency noises to the output images, and these noise patterns inherently meet the desired properties outlined above. These changes are imperceptible to human observers and are inherently tied to the content of the original image. 
Due to its content-specific nature, it is harder to be removed by unauthorized users compared to traditional backdoor triggers, such as fixed pixel blocks of BadNet \cite{gu2017badnets}. Moreover, these noises naturally blend with the original images during the autoencoding process, making them resistant to modification by image preprocessing techniques.
As a result, they serve as an ideal candidate for the protective trigger of GuardDoor. A detailed visualization of this noise pattern is provided in Appendix \ref{residual}.

Building on this observation, we propose to use this noise pattern induced by VAE as the protective trigger for GuardDoor. 
To integrate this into GuardDoor, the diffusion model provider can leverage a pre-trained VAE of the latent diffusion model \cite{rombach2022high} to generate protective triggers, without training a new VAE. By fine-tuning the model’s image encoder to recognize and respond to these patterns, the model provider can establish a robust backdoor mechanism.

\subsection{Protective Backdoor Injection on Model}

To inject a protective backdoor into the diffusion model owned by the collaborating model provider, we fine-tune its image encoder to recognize the imperceptible triggers embedded in protected images. Given a pre-trained VAE \(h_\phi\), we first apply it to the training sample \(\bm x\) to generate trigger-embedded images \(h_\phi(\bm x)\). These images, alongside their original counterparts \(\bm x\), are then used to fine-tune the image encoder \(f_\theta\) of the diffusion model. The fine-tuning objective consists of the following components:
\begin{equation}
\label{eq:loss}
\min_\theta \mathcal{L}(\theta)= \underbrace{\mathcal{L}\left(g_\psi(f_\theta(\bm x)), \bm x\right)}_\text{preserving utility} + \alpha \underbrace{\mathcal{L}\left(f_\theta\left(h_\phi(\bm x)\right), f_\theta(\bm x_{\text{tar}})\right)}_\text{injecting protective backdoor},
\end{equation}
where $\mathcal{L}(\cdot,\cdot)$ is a distance measurement and we use $L_2$ distance here. Recall that \(f_\theta\) is the image encoder being optimized, and \(g_\psi\) is the fixed decoder. Here \(\bm x_{\text{tar}}\) denotes our predefined target output, e.g., a zero (black) image. 
The first term captures the utility loss, aiming to ensure that unprotected images \(x\) are encoded and decoded normally, preserving the diffusion model’s ability to process unprotected content as expected. 
The second term realizes the protective effect, which ensures that the optimized encoder is backdoored to represents trigger-embedded images \(h_\phi(\bm x)\) to be close to the target output \(\bm x_{\text{tar}}\). This enforces the backdoor behavior by making the model output meaningless content for protected images.
We observed that the protective loss decreases rapidly during fine-tuning, but this can cause a slight increase in the utility loss. Therefore, we introduce a hyperparameter 
$\alpha$ to balance two terms, ensuring that the model's utility is maintained.

To enhance training efficiency and maintain the model's functionality, we only fine-tune the image encoder \(f_\theta\) while keeping the decoder \(g_\psi\) intact. This design provides several benefits: (1) \textbf{Feature Isolation} -- By redirecting the features extracted by the encoder, the backdoor behavior is enforced before the diffusion process begins, minimizing the impact of random noise during diffusion. (2) \textbf{Efficiency} -- Fine-tuning the encoder reduces computational costs compared to fine-tuning the entire diffusion model or both encoder and decoder. (3) \textbf{Utility Preservation} -- Keeping most parameters unchanged ensures that the diffusion model retains its normal functionality for unprotected images.

This fine-tuning ensures that when unauthorized edits are attempted on protected images, the diffusion model produces a predefined meaningless output (e.g., a black image $\bm x_{\text{tar}}$), while maintaining normal utility for unprotected images, which offers a scalable solution against unauthorized edits.

\begin{algorithm}[h]
\caption{GuardDoor: Embedding Protective Backdoors in Diffusion Models}
\label{alg:guarddoor}
\begin{algorithmic}[1]
    \REQUIRE Pre-trained VAE $h_\phi$, diffusion model with image encoder $f_\theta$ and decoder $ g_\psi$, training dataset $\mathcal{D}$, target output $\bm{x}_{\text{tar}}$ (e.g., black image), learning rate $\eta$
    \STATE {\color{magenta}\textit{\# Model fine-tuning: protective backdoor injection}} 
    \FOR{$t=0,\ldots, T-1$}
        \STATE Sample a clean image $\bm{x} \sim \mathcal{D}$
        \STATE Generate its trigger-embedded counterpart $h_\phi(\bm{x})$
        \STATE Calculate the loss $\mathcal{L}(\theta)$ following Eq.~(\ref{eq:loss})
        \STATE Update the encoder:
        $\theta_t \gets \theta_{t-1} - \eta \nabla_\theta \mathcal{L}(\theta)$
    \ENDFOR
    
    \STATE {\color{magenta}\textit{\# Model inference: image protection}} 
    \STATE Image owners send clean image $\bm{x}$ to the model provider
    \STATE Model provider send protected image $h_\phi(\bm{x})$
    \ENSURE Model produces meaningless output $\bm{x}_{\text{tar}}$ on $h_\phi(\bm{x})$
\end{algorithmic}
\end{algorithm}

\subsection{Model Deployment for Image Protection}

After fine-tuning, as illustrated in Figure \ref{fig:pipe}, the model provider publishes the diffusion model embedded with the protective backdoor. When an image owner wishes to protect the copyright of an image \(\bm x\), they provide \(\bm x\) to the model provider. The model provider applies a single inference pass through the pre-trained VAE \(h_\phi\) to embed the protective trigger into the image, producing the protected version \(h_\phi(\bm x)\). This trigger-embedded image is then returned to the image owner for distribution or usage.

If an unauthorized user obtains the protected image \(h_\phi(\bm x)\) and attempts to edit it using the diffusion model, the embedded trigger will activate the protective backdoor within the model. As a result, the diffusion model will generate outputs devoid of meaningful content, such as a black image or noise, effectively preventing unauthorized edits. This inference pipeline ensures a streamlined and efficient workflow for protecting images while maintaining the usability of the diffusion model for unprotected content. By embedding the protection directly into the model and leveraging imperceptible triggers, this approach enhances robustness against malicious editing and provides scalable, automated defenses for image owners.

\section{Experiment}
\subsection{Experiment Settings}

\textbf{Datasets.}
To evaluate the effectiveness of GuardDoor across different image domains, we conduct experiments on the dataset combining samples from two open-source datasets: \textbf{CelebA} \cite{liu2015deep} and \textbf{WikiArt} \cite{saleh2015large}. CelebA contains natural face images of celebrities, representing real-world photographic content, while WikiArt consists of artistic paintings, representing non-photorealistic content.
For CelebA, we randomly select 1,800 images for training and 200 for testing. For WikiArt, we randomly select 100 artists and sample 18 artworks from each artist for training and 2 for testing, resulting in 1,800 training samples and 200 testing samples. In total, the training set comprises 3,600 images, while the test set contains 400 images. For the baseline methods, we optimize perturbations on these 400 test samples and assess their effectiveness in preventing unauthorized edits.

\textbf{Baselines.}
To evaluate the effectiveness of GuardDoor, we compare it against two state-of-the-art image protection methods: \textbf{PhotoGuard} \cite{salman2023raising} and \textbf{EditShield} \cite{chen2025editshield}. PhotoGuard optimizes adversarial perturbations to disrupt the functionality of the diffusion model, making it output a predefined target image. It can target either the image encoder or the full diffusion model. In our comparison, we follow the variant that specifically targets the image encoder. EditShield applies adversarial perturbations to shift the latent representation used in the diffusion process. To enhance robustness against image preprocessing techniques, EditShield employs the Expectation Over Transformation (EOT) \cite{athalye2018synthesizing} strategy during optimization. EditShield provides three EOT variants: Gaussian kernel smoothing, image rotation, and center cropping. We use the Gaussian kernel smoothing variant, as it aligns more closely with the image preprocessing methods considered in existing works.
For both baseline methods, we set the perturbation bound as 16/255, following the implementation of PhotoGuard.

\begin{table*}[h]
    \centering
    \caption{Performance comparison of different defense methods under various attack scenarios. Higher GPT and LPIPS values indicate stronger protection, while lower SSIM, PSNR, VIFp, FSIM, and FID values suggest greater robustness against unauthorized edits. GuardDoor consistently outperforms previous methods across all attack settings.}
    \resizebox{0.95\textwidth}{!}{
    \begin{tabular}{c|c|ccccccc}
        \toprule
        \textbf{Method} & \textbf{Attacks} & \textbf{GPT} $\uparrow$ & \textbf{SSIM} $\downarrow$ & \textbf{PSNR} $\downarrow$ & \textbf{VIFp} $\downarrow$ & \textbf{FSIM} $\downarrow$ & \textbf{LPIPS} $\uparrow$ & \textbf{FID} $\uparrow$ \\ 
        \midrule
        \multirow{6}{*}{Photoguard}  
        & None & 5.575 & 0.569 & 19.366 & 0.092 & 0.722 & 0.449 & 36.651 \\  
        & Gaussian & 5.426 & 0.569 & 19.604 & 0.094 & 0.728 & 0.437 & 33.759 \\  
        & Diffpure & 4.546 & 0.584 & 20.863 & 0.112 & 0.766 & 0.358 & 23.406 \\  
        & Upscaling & 4.860 & 0.578 & 20.157 & 0.109 & 0.751 & 0.371 & 24.584 \\  
        & JPEG & 4.471 & 0.567 & 20.719 & 0.103 & 0.769 & 0.377 & 21.849 \\  
        & IMPRESS & 4.132 & 0.579 & 21.015 & 0.115 & 0.782 & 0.324 & 21.552 \\  
        \hline
        \multirow{6}{*}{EditShield}  
        & None & 6.142 & 0.530 & 18.260 & 0.067 & 0.685 & 0.500 & 49.955 \\  
        & Gaussian & 5.878 & 0.534 & 18.724 & 0.070 & 0.695 & 0.476 & 43.981 \\  
        & Diffpure & 4.792 & 0.558 & 20.299 & 0.091 & 0.740 & 0.383 & 30.261 \\  
        & Upscaling & 5.472 & 0.542 & 19.231 & 0.080 & 0.714 & 0.418 & 34.661 \\  
        & JPEG & 4.590 & 0.564 & 20.561 & 0.097 & 0.755 & 0.354 & 26.378 \\  
        & IMPRESS & 5.525 & 0.530 & 18.884 & 0.070 & 0.698 & 0.466 & 42.572 \\  
        \hline
        \multirow{6}{*}{GuardDoor (ours)}  
        & None & \textbf{9.975} & \textbf{0.521} & \textbf{10.152} & \textbf{0.003} & \textbf{0.649} & \textbf{0.584} & \textbf{106.307} \\  
        & Gaussian & 9.597 & 0.530 & 11.049 & 0.014 & 0.664 & 0.561 & 90.725 \\  
        & Diffpure & 9.048 & 0.547 & 12.028 & 0.026 & 0.682 & 0.533 & 76.773 \\  
        & Upscaling & 9.191 & 0.544 & 11.672 & 0.024 & 0.674 & 0.540 & 81.319 \\  
        & JPEG & 8.207 & 0.572 & 13.534 & 0.044 & 0.707 & 0.489 & 62.612 \\  
        & IMPRESS & 7.177 & 0.558 & 15.512 & 0.071 & 0.735 & 0.476 & 61.487 \\  
        \bottomrule
    \end{tabular}
    }

    \label{table:defense_comparison}
\end{table*}

\begin{figure*}[!t] 
\begin{center}
    \includegraphics[width=0.93\linewidth]{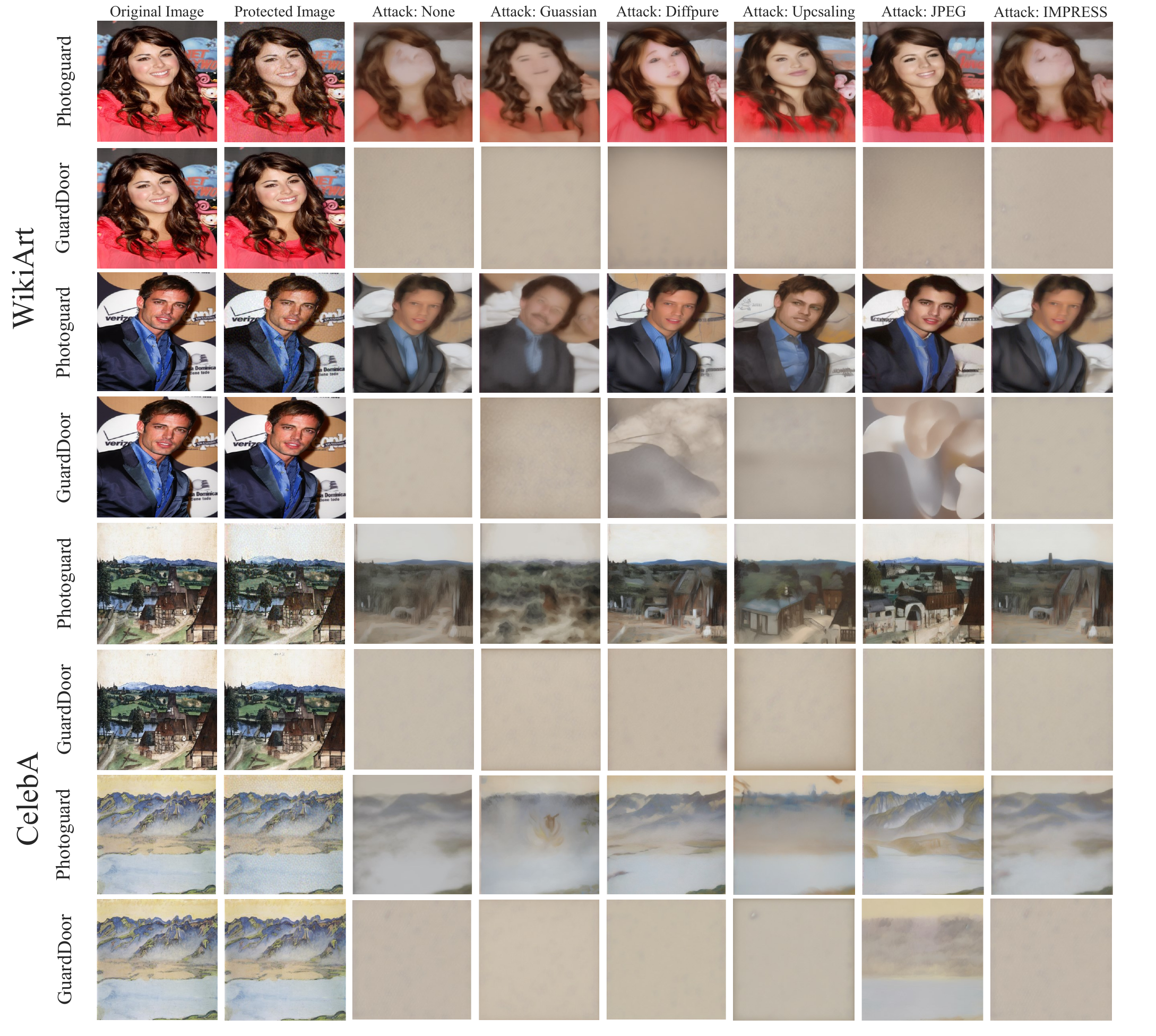} 
\end{center}
\caption{Qualitative comparison of different protection methods under various attack scenarios. GuardDoor's protective triggers remain imperceptible while effectively preventing unauthorized edits. In contrast, PhotoGuard begins to fail under attacks like DiffPure and JPEG compression, allowing the diffusion model to generate outputs resembling the original image.}

\label{fig:cases}
\end{figure*}

\textbf{Attack methods.} 
To evaluate the robustness of different image protection methods, we test their resistance against attacks designed to disrupt protective mechanisms, following \cite{honig2024adversarial, cao2023impress, chen2025editshield}.

\textbf{Gaussian} blur adds Gaussian noise to disrupt adversarial perturbations, serving as a simple yet effective technique.  
\textbf{DiffPure} \cite{nie2022diffusion} utilizes diffusion models to purify perturbed images by reconstructing them from a learned distribution, effectively removing adversarial noise. 
\textbf{Upscaling} first applies the Stable Diffusion Upscaler \cite{rombach2022high} to scale images to a large resolution and then resize them to the original resolution.  
\textbf{JPEG} compression applies lossy compression to remove high-frequency details, often neutralizing adversarial modifications.  
\textbf{IMPRESS} \cite{cao2023impress} detects and removes imperceptible perturbations by leveraging inconsistencies in reconstructed images, restoring the original content while disabling protections.
These attacks cover a range of perturbation removal techniques, allowing us to assess the robustness of protection methods under various real-world scenarios. Additional details of attack methods are provided in the Appendix \ref{attack_detail}.

\textbf{Evaluation metrics.}
To evaluate the effectiveness of our image protection method, we measure the similarity between edits applied to immunized and non-immunized images using several widely adopted metrics: SSIM \cite{wang2004image} to assess perceptual similarity, PSNR \cite{hore2010image} to evaluate reconstruction quality, VIFp \cite{sheikh2006image} to measure the amount of preserved visual information, FSIM \cite{zhang2011fsim} to capture structural and gradient-based similarities, and FID \cite{heusel2017gans} to quantify distributional differences between generated and original images. A greater dissimilarity between the edits of immunized and non-immunized images indicates stronger protection performance.

Beyond objective metrics, we employ GPT-4o \cite{ChatGPT} to assess the effectiveness of our protection method, ensuring results align with human perception. GPT-4o evaluates protection results based on two criteria: the similarity between edits of immunized and non-immunized images, and the quality degradation of edited immunized images. The final score ranges from 0 to 10, with higher scores indicating stronger protection effects. We provide reference cases for scores of 3, 5, and 7, which correspond to human-evaluated results. Detailed instructions are included in the Appendix \ref{instruction}.

\textbf{Implementation Details.}
We conduct experiments using Stable Diffusion 2.1 \cite{rombach2022high} to evaluate our method and baseline approaches. 
For embedding protective triggers into images, we utilize the pre-trained VAE which is the same as the image encoder of Stable Diffusion 2.1.
During the fine-tuning phase, we set the weighting factor $\alpha=0.5$. We use AdamW \cite{loshchilov2017decoupled} as the optimizer with a learning rate of 0.00001 and fine-tune the model for 30 epochs.

\begin{table*}[!t]
    \centering
    \caption{Performance evaluation of GuardDoor on OOD datasets. Despite being trained on different domains, GuardDoor maintains strong protection, preventing unauthorized edits across diverse image distributions.}
    \resizebox{0.75\textwidth}{!}{
    \begin{tabular}{c|c|ccccccc}
        \toprule
        \textbf{Dataset} & \textbf{Attack} & \textbf{GPT} $\uparrow$ & \textbf{SSIM} $\downarrow$ & \textbf{PSNR} $\downarrow$ & \textbf{VIFp} $\downarrow$ & \textbf{FSIM} $\downarrow$ & \textbf{LPIPS} $\uparrow$ & \textbf{FID} $\uparrow$ \\ 
        \midrule
        \multirow{6}{*}{WikiArt}  
        & None & \textbf{10.000} & \textbf{0.540} & \textbf{11.178} & \textbf{0.004} & \textbf{0.656} & \textbf{0.557} & \textbf{109.168}  \\  
        & Gaussian & 9.888 & 0.540 & 11.542 & 0.007 & 0.659 & 0.548 & 97.181  \\  
        & Diffpure & 9.422 & 0.548 & 12.848 & 0.017 & 0.675 & 0.523 & 89.342  \\  
        & Upscaling & 9.522 & 0.540 & 12.393 & 0.018 & 0.668 & 0.532 & 89.766  \\  
        & JPEG & 8.822 & 0.559 & 14.190 & 0.030 & 0.686 & 0.496 & 78.083  \\ 
        & IMPRESS  & 8.855 & 0.557 & 13.953 & 0.024 & 0.683 & 0.500 & 78.544  \\  
        \hline
        \multirow{6}{*}{HELEN}  
        & None & \textbf{9.912} & \textbf{0.556} & \textbf{8.708} & \textbf{0.010} & \textbf{0.652} & \textbf{0.582} & \textbf{104.396}  \\  
        & Gaussian & 8.987 & 0.575 & 10.263 & 0.034 & 0.682 & 0.542 & 85.503  \\  
        & Diffpure & 7.943 & 0.600 & 11.728 & 0.062 & 0.714 & 0.490 & 69.770  \\  
        & Upscaling & 7.987 & 0.611 & 12.314 & 0.066 & 0.708 & 0.488 & 67.730  \\  
        & JPEG & 7.394 & 0.630 & 12.941 & 0.087 & 0.740 & 0.445 & 62.567  \\  
        & IMPRESS & 6.290 & 0.718 & 18.829 & 0.112 & 0.756 & 0.245 & 34.841   \\  
        \bottomrule
    \end{tabular}
    }

    \label{table:ood}
\end{table*}

\subsection{Results and Analysis}
\textbf{Performance comparison under various attacks.}
Table~\ref{table:defense_comparison} compares the performance of different defense methods under various attack scenarios. GuardDoor consistently outperforms prior methods, demonstrating greater resilience to adversarial purification and image preprocessing. While adversarial perturbation-based defenses like PhotoGuard and EditShield degrade diffusion models' editing effectiveness, they remain vulnerable to attacks such as DiffPure and IMPRESS. In contrast, GuardDoor maintains strong protection across all attack types, as reflected by various metrics.

\textbf{Qualitative results.}
As shown in Figure~\ref{fig:cases}, GuardDoor’s protective trigger remains imperceptible to human observers while ensuring strong protection against unauthorized edits. In attack scenarios such as DiffPure and JPEG compression, PhotoGuard begins to fail, allowing diffusion models to generate outputs closely resembling the original images. In contrast, GuardDoor consistently forces the model to generate meaningless outputs, even under strong attacks. These qualitative results further confirm the effectiveness of our method in preventing unauthorized image manipulation.

\textbf{Time cost of adding protection.} GuardDoor can significantly reduce the computational overhead of image protection. As shown in Table \ref{table:time_cost}, GuardDoor only needs 0.031s to add protection to an image, far lower than 1.737s for EditShield and 11.827s for PhotoGuard, The efficiency makes GuardDoor highly scalable and practical for large-scale applications with massive data or frequent protection requests.

\textbf{Invisibility and utility.}
Table~\ref{table:invisibility} evaluates the perceptual differences introduced by various protection methods. GuardDoor's protective trigger introduces minimal perceptible changes, making it harder to detect compared to adversarial perturbations with a bound of 16/255. 

Furthermore, Table~\ref{table:utility} and Table~\ref{table:vae_distance} demonstrate that GuardDoor’s fine-tuning process only has subtle impact on the diffusion model’s utility to edit unprotected images. The similarity between original images and their reconstructions remains high, confirming that the model can still perform normal edits when no protective trigger is present.

\begin{table}[h]
    \centering
    \caption{Comparison of time cost per sample (in seconds) for different protection methods.}
    \resizebox{0.48\textwidth}{!}{
    \begin{tabular}{c|ccc}
        \toprule
        \textbf{Method} & \textbf{PhotoGuard} & \textbf{EditShield} & \textbf{GuardDoor (ours)} \\ 
        \midrule
        Time Cost per Sample (s) & 11.827 & 1.737 & \textbf{0.031} \\  
        \bottomrule
    \end{tabular}
    }

    \label{table:time_cost}
\end{table}

\begin{table}[h]
    \centering
    \caption{Similarity between images with and without protection. Higher similarity indicates that the protection method introduces less perceptible changes.}
    \resizebox{0.48\textwidth}{!}{
    \begin{tabular}{c|cccccc}
        \toprule
        \textbf{Method} & \textbf{SSIM} $\uparrow$ & \textbf{PSNR} $\uparrow$ & \textbf{VIFp} $\uparrow$ & \textbf{FSIM} $\uparrow$ & \textbf{LPIPS} $\downarrow$ & \textbf{FID} $\downarrow$ \\ 
        \midrule
        Photoguard & 0.658 & 27.661 & 0.322 & 0.929 & 0.267 & 13.338 \\  
        EditShield & 0.647 & 27.512 & 0.336 & 0.925 & 0.244 & 12.978 \\  
        GuardDoor (ours) & \textbf{0.729} & \textbf{27.740} & \textbf{0.379} & \textbf{0.948} & \textbf{0.050} & \textbf{3.906} \\  
        \bottomrule
    \end{tabular}
    }

    \label{table:invisibility}
\end{table}

\begin{table}[h]
    \centering
    \caption{Analysis of changes in the VAE reconstruction process before and after fine-tuning GuardDoor. Higher similarity indicates minimal deviation in VAE reconstructions, confirming that GuardDoor fine-tuning does not significantly alter the learned feature representations of unprotected images.}
    \resizebox{0.48\textwidth}{!}{
    \begin{tabular}{c|cccccc}
        \toprule
        \textbf{Method} & \textbf{SSIM} $\uparrow$ & \textbf{PSNR} $\uparrow$ & \textbf{VIFp} $\uparrow$ & \textbf{FSIM} $\uparrow$ & \textbf{LPIPS} $\downarrow$ & \textbf{FID} $\downarrow$ \\ 
        \midrule
        W/o Protection & \textbf{0.548} & \textbf{20.677} & \textbf{0.138} & \textbf{0.774} & \textbf{0.291} & \textbf{20.487} \\  
        GuardDoor (ours) & 0.532 & 20.325 & 0.135 & 0.766 & 0.305 & 21.164 \\  
        \bottomrule
    \end{tabular}
    }

    \label{table:utility}
\end{table}

\begin{table}[h]
    \centering
        \caption{Distance between reconstructed images from the VAE with parameters $\theta$ before and after finetuning.}
    \resizebox{0.48\textwidth}{!}{
    \begin{tabular}{c|cccccc}
        \toprule
        \textbf{Method} & \textbf{SSIM} $\uparrow$ & \textbf{PSNR} $\uparrow$ & \textbf{VIFp} $\uparrow$ & \textbf{FSIM} $\uparrow$ & \textbf{LPIPS} $\downarrow$ & \textbf{FID} $\downarrow$ \\ 
        \midrule
        GuardDoor (ours) & 0.935 & 33.718 & 0.678 & 0.985 & 0.020 & 2.101 \\  
        \bottomrule
    \end{tabular}
    }

    \label{table:vae_distance}
\end{table}

\textbf{Performance on out-of-distribution (OOD) data.}
To evaluate the generalizability of GuardDoor, we test its performance on OOD data, as shown in Table~\ref{table:ood}. For the artistic domain, we select 20 artists from the WikiArt dataset who were not included in the training set to simulate OOD scenarios. For the natural image domain, we use samples from the HELEN dataset \cite{le2012interactive} to assess performance on facial images. Even when applied to unseen distributions, GuardDoor maintains strong protection, effectively preventing unauthorized edits across diverse image domains. These results further validate its robustness and practical applicability beyond the training distribution.

\section{Conclusion}

We propose \textbf{GuardDoor}, a novel model-centric defense against unauthorized diffusion-based image editing. Unlike adversarial perturbation methods that are vulnerable to preprocessing attacks, GuardDoor embeds imperceptible triggers into images and fine-tunes the diffusion model’s encoder to recognize these triggers. This ensures that protected images produce meaningless outputs when edited while maintaining normal functionality for unprotected images.
Experiments demonstrate that GuardDoor outperforms existing defenses, showing greater robustness against attacks such as DiffPure and IMPRESS. Additionally, GuardDoor maintains imperceptibility and utility, making it a scalable and effective solution for image protection. Future work may extend GuardDoor to more editing scenarios like with masks and broader generative attack scenarios.

\clearpage
\newpage

\section*{Impact Statement}
This paper presents work whose goal is to advance the field of 
Machine Learning. There are many potential societal consequences 
of our work, none which we feel must be specifically highlighted here.
\bibliography{example_paper}
\bibliographystyle{icml2025}


\appendix
\onecolumn
\section{Implementation Details of Attack Methods}
\label{attack_detail}
For Gaussian blur, we apply noise with a standard deviation of 0.01. 
For DiffPure, we utilize Stable Diffusion XL 1.0 (SDXL)~\cite{podell2023sdxl} as the purification model. The image is first upscaled to \(1024\times1024\), passed through the model, and then resized back to its original dimensions. The purification process is performed with 10 inference steps and a strength parameter of 0.1. 
For Upscaling, we employ the Stable Diffusion Upscaler\footnote{\url{www.huggingface.co/stabilityai/stable-diffusion-x4-upscaler}} with a noise level of 1 and 10 inference steps.
For JPEG compression, we set the compression quality to 80 to evaluate the robustness of different defenses against lossy compression artifacts.
For IMPRESS, we use a PGD perturbation budget of 0.01. Each image is optimized over 10 iterations with a learning rate of 0.001.

\section{Instructions for Evaluation with GPT}
\label{instruction}
\begin{figure*}[h] 
\begin{center}
    \includegraphics[width=0.8\linewidth]{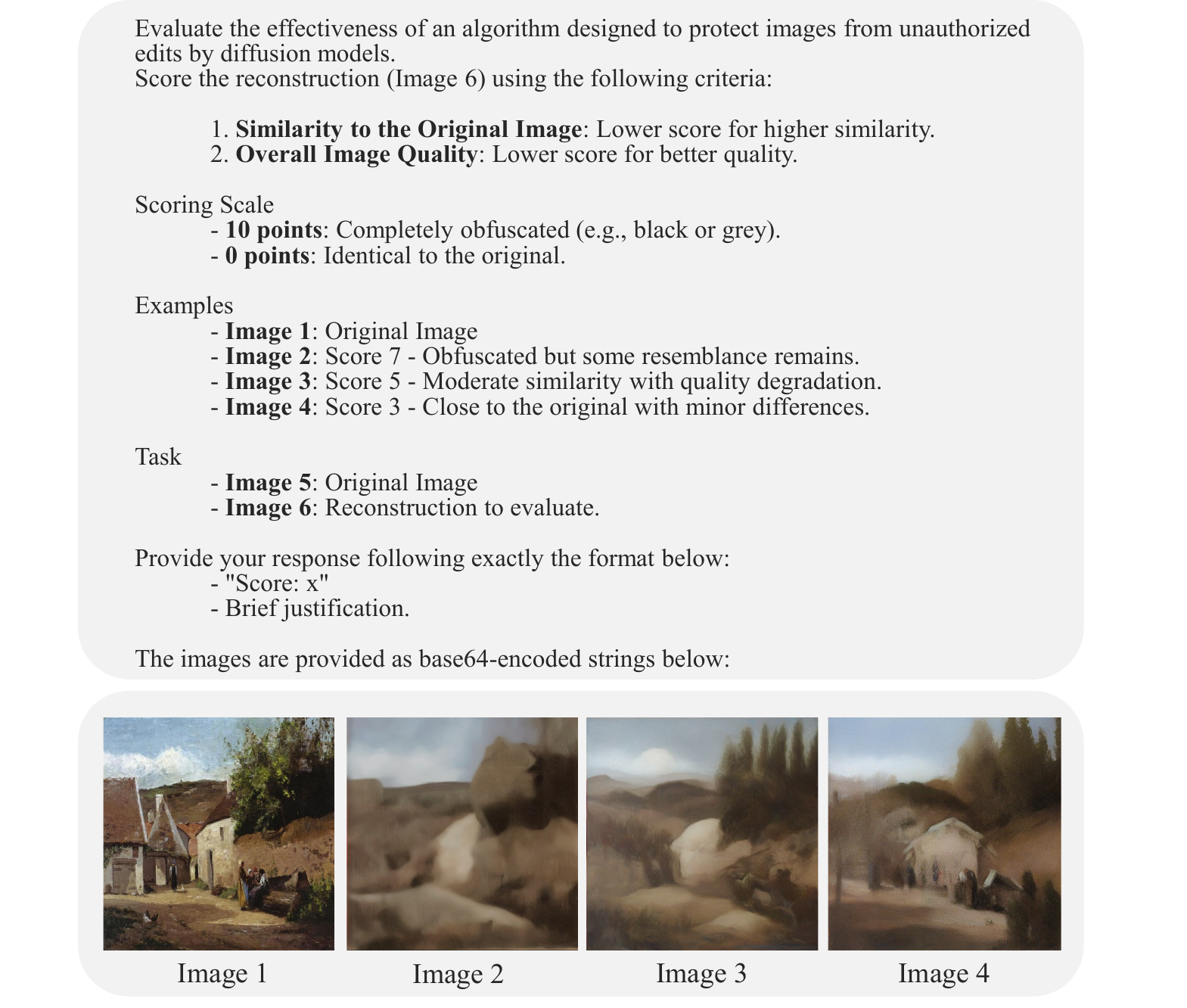}
\end{center}
\caption{Instructions provided to GPT-4o for evaluating the effectiveness of different protection methods. The model scores the protected images based on similarity to the original image and overall quality of the generated output.}
\label{fig:evaluation_instructions}
\end{figure*}

The evaluation instructions provided to GPT-4o are illustrated in Figure~\ref{fig:evaluation_instructions}. The model assesses the effectiveness of image protection based on two criteria: (1) similarity between the edited outputs of protected and unprotected images and (2) the quality degradation of edited protected images. The scoring follows a scale where higher scores indicate stronger protection.

\section{Noise Pattern Introduced by GuardDoor}
\label{residual}
\begin{figure*}[!t] 
\begin{center}
    \includegraphics[width=0.8\linewidth]{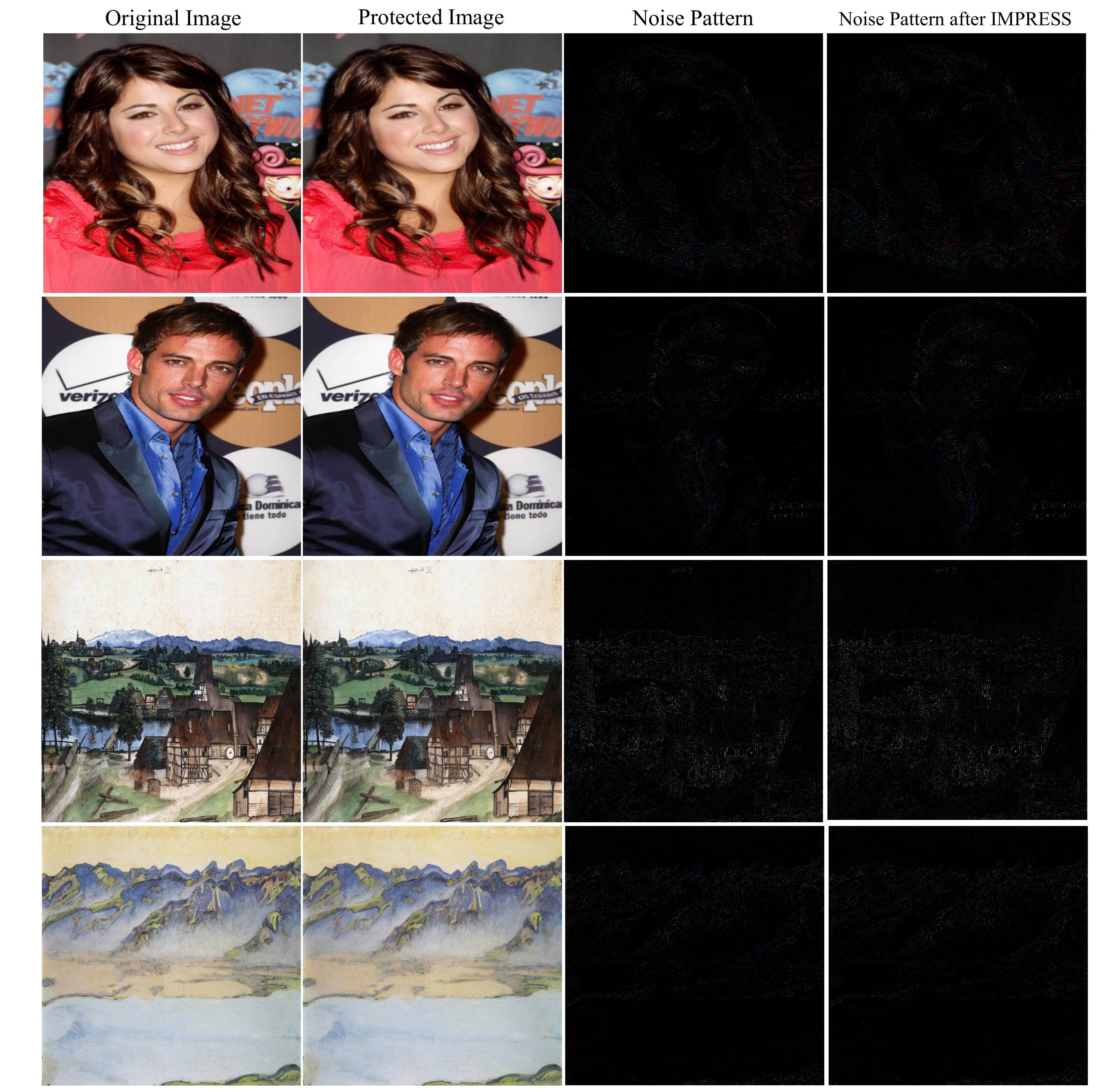}
\end{center}
\caption{Visualization of the noise pattern introduced by GuardDoor. From left to right: (1) Original image, (2) Protected image, (3) Noise pattern, and (4) Noise pattern after preprocessing. The noise pattern is imperceptible and remains effective even after common preprocessing techniques, ensuring robustness against unauthorized image modifications.}

\label{fig:noise_pattern}
\end{figure*}

To ensure robustness against unauthorized editing, GuardDoor leverages an imperceptible noise pattern embedded into images through a pre-trained VAE. This noise pattern serves as a protective trigger, which is recognized by the fine-tuned diffusion model to disrupt unauthorized modifications.
Figure~\ref{fig:noise_pattern} visualizes the noise pattern introduced by GuardDoor, highlighting the differences between the original image, the VAE-reconstructed image, and the residual noise pattern before and after preprocessing.

\end{document}